\documentclass[english]{article}
\usepackage[latin9]{inputenc}
\usepackage{geometry}
\usepackage{color}
\usepackage{amsmath}
\usepackage{amssymb}
\usepackage{graphicx}
\usepackage{esint}

\makeatletter

\providecommand{\tabularnewline}{\\}

\@ifundefined{date}{}{\date{}}

\usepackage{array}\usepackage{float}
\usepackage{amsbsy}\usepackage{babel}

\makeatother

\usepackage{babel}
\begin{document}

\title{Cumulative Restricted Boltzmann Machines for Ordinal Matrix Data
Analysis}

\author{Truyen Tran$^{\dagger\ddagger}$, Dinh Phung$^{\dagger}$, Svetha
Venkatesh$^{\dagger}$ \\
 $^{\dagger}$Pattern Recognition and Data Analytics, Deakin University,
Australia\\
 $^{\ddagger}$Department of Computing, Curtin University, Australia\\
 \{truyen.tran,dinh.phung,svetha.venkatesh\}@deakin.edu.au }
\maketitle
\begin{abstract}
Ordinal data is omnipresent in almost all multiuser-generated feedback
- questionnaires, preferences etc. This paper investigates modelling
of ordinal data with Gaussian restricted Boltzmann machines (RBMs).
In particular, we present the model architecture, learning and inference
procedures for both vector-variate and matrix-variate ordinal data.
We show that our model is able to capture latent opinion profile of
citizens around the world, and is competitive against state-of-art
collaborative filtering techniques on large-scale public datasets.
The model thus has the potential to extend application of RBMs to
diverse domains such as recommendation systems, product reviews and
expert assessments. 
\end{abstract}
\global\long\def\BigO{\mathcal{O}}
 \global\long\def\U{\mathcal{U}}
 \global\long\def\I{\mathcal{I}}
 \global\long\def\M{\mathcal{M}}
 \global\long\def\B{\mathcal{B}}
 \global\long\def\Loss{\mathcal{L}}
 \global\long\def\userset{\mathcal{U}}
 \global\long\def\itemset{\mathcal{I}}
 \global\long\def\ratemat{\mathcal{R}}
 \global\long\def\A{\mathcal{A}}

\global\long\def\rateset{\mathcal{S}}
 \global\long\def\hb{\boldsymbol{h}}
 \global\long\def\h{h}
 \global\long\def\vb{\boldsymbol{v}}
 \global\long\def\v{v}
 \global\long\def\xb{\boldsymbol{u}}
 \global\long\def\x{u}
 \global\long\def\ub{\boldsymbol{u}}
 \global\long\def\ubb{\mathfrak{u}}
 \global\long\def\u{u}

\global\long\def\wb{\boldsymbol{w}}
 \global\long\def\w{w}
 \global\long\def\qb{\boldsymbol{q}}
 \global\long\def\q{q}
 \global\long\def\pb{\boldsymbol{p}}
 \global\long\def\p{p}
 \global\long\def\gb{\boldsymbol{g}}
 \global\long\def\Hb{\boldsymbol{H}}
 \global\long\def\Vb{\boldsymbol{V}}

\global\long\def\thetab{\boldsymbol{\theta}}
 \global\long\def\alphab{\boldsymbol{\alpha}}
 \global\long\def\etab{\boldsymbol{\eta}}
 \global\long\def\nub{\boldsymbol{\nu}}
 \global\long\def\betab{\boldsymbol{\beta}}
 \global\long\def\mub{\boldsymbol{\mu}}
 \global\long\def\gammab{\boldsymbol{\gamma}}

\global\long\def\taub{\boldsymbol{\tau}}
 \global\long\def\lambdab{\boldsymbol{\lambda}}
 \global\long\def\LL{\mathcal{L}}
 \global\long\def\Domain{\Omega}
 \global\long\def\Domainb{\boldsymbol{\Omega}}

\global\long\def\Real{\mathbb{R}}
 \global\long\def\Id{\mathbb{I}}
 \global\long\def\Normal{\mathcal{N}}
 \global\long\def\Model{\mathrm{CRBM}}

\section{Introduction}

Restricted Boltzmann machines (RBMs) \cite{smolensky1986information,freund1994unsupervised,le2008representational}
have recently attracted significant interest due to their versatility
in a variety of unsupervised and supervised learning tasks \cite{Salakhutdinov-et-alICML07,larochelle2008classification,mohamed2010phone},
and in building deep architectures \cite{hinton2006rdd,salakhutdinov2009deep}.
A RBM is a bipartite undirected model that captures the generative
process in which a data vector is generated from a binary hidden vector.
The bipartite architecture enables very fast data encoding and sampling-based
inference; and together with recent advances in learning procedures,
we can now process massive data with large models \cite{Hinton02,tieleman2008training,marlin2010inductive}.

This paper presents our contributions in developing RBM specifications
as well as learning and inference procedures for multivariate ordinal
data. This extends and consolidates the reach of RBMs to a wide range
of user-generated domains - social responses, recommender systems,
product/paper reviews, and expert assessments of health and ecosystems
indicators. Ordinal variables are qualitative in nature -- the absolute
numerical assignments are not important but the relative order is.
This renders numerical transforms and real-valued treatments inadequate.
Current RBM-based treatments, on the other hand, ignore the ordinal
nature and treat data as unordered categories \cite{Salakhutdinov-et-alICML07,Truyen:2009a}.
While convenient, this has several drawbacks: First, order information
is not utilised, leading to more parameters than necessary - each
category needs parameters. Second, since categories are considered
independently, it is less interpretable in terms of how ordinal levels
are generated. Better modelling should account for the ordinal generation
process.

Adapting the classic idea from \cite{mccullagh1980rmo}, we assume
that each ordinal variable is generated by an underlying latent utility,
along with a threshold per ordinal level. As soon as the utility passes
the threshold, its corresponding level is selected. As a result, this
process would implicitly encode the order. Our main contribution here
is a novel RBM architecture that accounts for multivariate, ordinal
data. More specifically, we further assume that the latent utilities
are Gaussian variables connected to a set of binary hidden factors
(i.e., together they form a Gaussian RBM \cite{hinton2006rdd}). This
offers many advantages over the standard approach that imposes a fully
connected Gaussian random field over utilities \cite{kottas2005nonparametric,jeliazkov2008fitting}:
First, utilities are seen as being generated from a set of binary
factors, which in many cases represent the user's hidden profile.
Second, utilities are decoupled given the hidden factors, making parallel
sampling easier. And third, the posteriors of binary factors can be
estimated from the ordinal observations, facilitating dimensionality
reduction and visualisation. We term our model Cumulative RBM ($\Model$)%
\footnote{The term 'cumulative' is to be consistent with the statistical literature
when referring to the ordinal treatment in \cite{mccullagh1980rmo}.%
}.

This new model behaves differently from standard Gaussian RBMs since
utilities are never observed in full. Rather, when an ordinal level
of an input variable is observed, it poses an \emph{interval constraint}
over the corresponding utility. The distribution over the utilities
now becomes a \emph{truncated} multivariate Gaussian. This also has
another consequence during learning: While in standard RBMs we need
to sample for the \emph{free-phase} only (e.g., see \cite{Hinton02}),
now we also need to sample for the \emph{clamped-phase. }As a result,
we introduce a double persistent contrastive divergence (PCD) learning
procedure, as opposed to the single PCD in \cite{tieleman2008training}.

The second contribution is in advancing these ordinal RBMs from modelling
i.i.d. vectors to modelling matrices of correlated entries. These
ordinal matrices are popular in multiuser-generated assessments: Each
user would typically judge a number of items producing a user-specific
data vector where \emph{intra-vector} entries are inherently correlated.
Since user's choices are influenced by their peers, these \emph{inter-vector}
entries are no longer independent. The idea is borrowed from a recent
work in \cite{Truyen:2009a} which models both the user-specific and
item-specific processes. More specifically, an ordinal entry is assumed
to be jointly generated from user-specific latent factors and item-specific
latent factors. This departs significantly from the standard RBM architecture:
we no longer map from a visible vector to an hidden vector but rather
map from a visible matrix to two hidden matrices.

In experiments, we demonstrate that our proposed $\Model$ is capable
of capturing the latent profile of citizens around the world. Our
model is also competitive against state-of-the-art collaborative filtering
methods on large-scale public datasets.

We start with the RBM structure for ordinal vectors in Section~\ref{sec:CRBM-for-Vectors},
and end with the general structure for ordinal matrices in Section~\ref{sec:CBM-for-Matrix}.
Section~\ref{sec:Experiments} presents experiments validating our
ordinal RBMs in modelling citizen's opinions worldwide and in collaborative
filtering. Section~\ref{sec:Related-Work} discusses related work,
which is then followed by the conclusions.

\section{Cumulative RBM for Vectorial Data \label{sec:CRBM-for-Vectors}}

\subsection{Model Definition \label{sub:Model-Structure-and-Param}}

\begin{figure}
\begin{centering}
\includegraphics[width=0.3\textwidth]{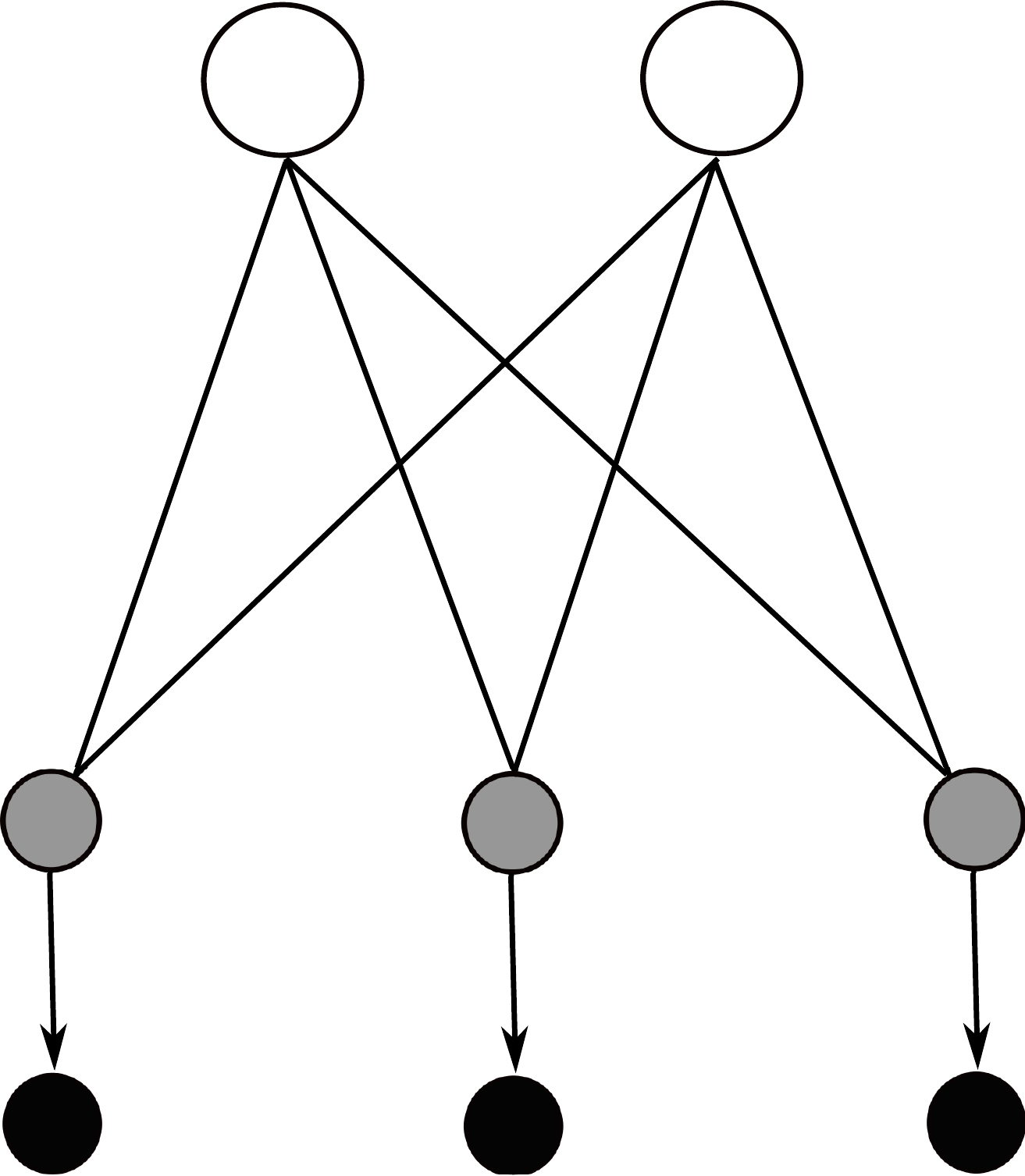} 
\par\end{centering}

\caption{Model architecture of the Cumulative Restricted Boltzmann Machine
($\Model$). Filled nodes represent observed ordinal variables, shaded
nodes are Gaussian utilities, and empty nodes represent binary hidden
factors. \label{fig:Model-architecture.}}
\end{figure}

Denote by $\vb=(\v_{1},\v_{2},...,\v_{N})$ the set of ordinal observations.
For ease of presentation we assume for the moment that observations
are homogeneous, i.e., observations are drawn from the same discrete
ordered category set $S=\{c_{1}\prec c_{2}\prec...,\prec c_{L}\}$
where $\prec$ denotes the order in some sense. We further assume
that each ordinal $v_{i}$ is solely generated from an underlying
latent \emph{utility} $u_{i}\in\Real$ as follows \cite{mccullagh1980rmo}
\begin{equation}
P(v_{i}=c_{l}\mid u_{i})=\begin{cases}
\mathbb{I}\left[-\infty<\x_{i}\le\theta_{i1}\right] & l=1\\
\mathbb{I}\left[\theta_{i(l-1)}<u_{i}\le\theta_{il}\right] & 1<l\le L-1\\
\mathbb{I}\left[\theta_{i(L-1)}<u_{i}<\infty\right] & l=L
\end{cases}\label{eq:ordinal-model}
\end{equation}
where $\theta_{i1}<\theta_{i2}<...<\theta_{i(L-1)}$ are threshold
parameters. In words, we choose an ordered category on the basis of
the interval to which the underlying utility belongs.

The utilities are connected with a set of hidden \emph{binary factors}
$\hb=(h_{1},h_{2},...,h_{K})\in\left\{ 0,1\right\} ^{K}$ so that
the two layers of $(\ub,\hb)$ form a undirectional bipartite graph
of Restricted Boltzmann Machines (RBMs) \cite{smolensky1986information,freund1994unsupervised,hinton2006rdd}.
Binary factors can be considered as the hidden features that govern
the generation of the observed ordinal data. Thus the generative story
is: we start from the binary factors to generate utilities, which,
in turn, generate ordinal observations. See, for example, Fig.~\ref{fig:Model-architecture.}
for a graphical representation of the model.

Let $\Psi(\ub,\hb)\ge0$ be the model potential function, which can
be factorised as a result of the bipartite structure as follows 
\[
\Psi(\ub,\hb)=\left[\prod_{i}\phi_{i}(\x_{i})\right]\left[\prod_{i,k}\psi_{ik}(\x_{i},h_{k})\right]\left[\prod_{k}\phi_{k}(h_{k})\right]
\]
where $\phi_{i},\psi_{ik}$ and $\phi_{k}$ are local potential functions.
The model joint distribution is defined as

\begin{equation}
P(\vb,\xb,\hb)=\frac{1}{Z}\Psi(\ub,\hb)\prod_{i}P(v_{i}\mid\x_{i})\label{eq:Model-def}
\end{equation}
where $Z=\int_{\ub}\sum_{\hb}\Psi(\ub,\hb)d\ub$ is the normalising
constant.

We assume the utility layer and the binary factor layer form a Gaussian
RBM%
\footnote{This is for convenience only. In fact, we can replace Gaussian by
any continuous distribution in the exponential family.%
} \cite{hinton2006rdd}. This translates into the local potential functions
as follows

\begin{eqnarray}
 & \phi_{i}(\x_{i})=\exp\left\{ -\frac{\x_{i}^{2}}{2\sigma_{i}^{2}}+\alpha_{i}\x_{i}\right\} ;\quad\psi_{ik}(\x_{i},h_{k})=\exp\left\{ w_{ik}u_{i}h_{k}\right\} ;\quad\phi_{k}(h_{k})=\exp\left\{ \gamma_{k}h_{k}\right\} \label{eq:poten-params}
\end{eqnarray}
where $\sigma_{i}$ is the standard deviation of the $i$-th utility,
$\left\{ \alpha_{i},\gamma_{k},w_{ik}\right\} $ are free parameters
for $i=1,2,..,N$ and $k=1,2,..,K$.

The ordinal assumption in Eq.~(\ref{eq:ordinal-model}) introduces
\emph{hard constraints} that we do not see in standard Gaussian RBMs.
Whenever an ordered category $\v_{i}$ is observed, the corresponding
utility is automatically \emph{truncated}, i.e., $u_{i}\in\Omega(v_{i})$,
where $\Omega(v_{i})$ is the new domain of $u_{i}$ defined by $v_{i}$
as in Eq.~(\ref{eq:ordinal-model}). In particular, the utility is
truncated from above if the ordinal level is the lowest, from below
if the level is the largest, and from both sides otherwise. For example,
the conditional distribution of the\emph{ }latent utility $P(\x_{i}\mid v_{i},\hb)$
is a truncated Gaussian

\begin{eqnarray}
P(\x_{i}\mid v_{i},\hb) & \propto & \mathbb{I}\left[\u_{i}\in\Omega(v_{i})\right]\mathcal{N}\left(\x_{i};\mu_{i}(\hb),\sigma_{i}\right)\label{eq:cond-utility-Gauss}
\end{eqnarray}
where $\mathcal{N}\left(\x_{i};\mu_{i}(\hb),\sigma_{i}\right)$ is
the normal density distribution of mean $\mu_{i}(\hb)$ and standard
deviation $\sigma_{i}$. The mean $\mu_{i}(\hb)$ is computed as 
\begin{equation}
\mu_{i}(\hb)=\sigma_{i}^{2}\left(\alpha_{i}+\sum_{k=1}^{K}w_{ik}h_{k}\right)\label{eq:mean-structure}
\end{equation}

As a generative model, we can estimate the probability that an ordinal
level is being generated from hidden factors $\hb$ as follows 
\begin{eqnarray}
P(v_{i}=c_{l}\mid\hb) & = & \int_{u_{i}\in\Omega(c_{l})}P(u_{i}|\hb)=\begin{cases}
\Phi\left(\theta_{1}^{*}\right) & l=1\\
\Phi\left(\theta_{l}^{*}\right)-\Phi\left(\theta_{(l-1)}^{*}\right) & 1<l\le L-1\\
1-\Phi(\theta_{L-1}^{*}) & l=L
\end{cases}\label{eq:ordinal-gen-given-rep}
\end{eqnarray}
where $\theta_{l}^{*}=\frac{\theta_{l}-\mu_{i}(\hb)}{\sigma_{i}}$,
and $\Phi(\cdot)$ is the cumulative distribution function of the
Gaussian. Given this property, we term our model by Cumulative Restricted
Boltzmann Machine ($\Model$).

Finally, the thresholds are parameterised so that the lowest threshold
is fixed to a constant $\theta_{i1}=\tau_{i1}$ and the higher thresholds
are spaced as $\theta_{il}=\theta_{i(l-1)}+e^{\tau_{il}}$ with free
parameter $\tau_{il}$ for $l=2,3,..,L-1$.

\subsection{Factor Posteriors \label{sub:Factor-Posteriors}}

Often we are interested in the posterior of factors $\left\{ P(h_{k}\mid\vb)\right\} _{k=1}^{K}$
as it can be considered as a summary of the data $\vb$. The nice
thing is that it is now numerical and can be used for other tasks
such as clustering, visualisation and prediction.

Like standard RBMs, the factor posteriors given the utilities are
conditionally independent and assume the form of logistic units

\begin{equation}
P(h_{k}=1\mid\ub)=\frac{1}{1+\exp\left(-\gamma_{k}-\sum_{i}w_{ik}u_{i}\right)}\label{eq:posterior}
\end{equation}
However, since the utilities are themselves hidden, the posteriors
given only the ordinal observations are not independent: 
\begin{equation}
P(h_{k}\mid\vb)=\sum_{\hb_{\neg k}}\int_{\ub\in\Domainb(\vb)}P(\hb,\ub|\vb)d\ub\label{eq:data-representation}
\end{equation}
where $\hb_{\neg k}=\hb\backslash h_{k}$ and $\Domainb(\vb)=\Domain(v_{1})\times\Domain(v_{2})\times...\Domain(v_{N})$
is the domain of the utility constrained by $\vb$ (see Eq.~(\ref{eq:ordinal-model})).
Here we describe two approximation methods, namely Markov chain Monte
Carlo (MCMC) and variational method (mean-field).

\paragraph{MCMC.}

We can exploit the bipartite structure of the RBM to run layer-wise
Gibbs sampling: sample the truncated utilities in parallel using Eq.~(\ref{eq:cond-utility-Gauss})
and the binary factors using Eq.~(\ref{eq:posterior}). Finally,
the posteriors are estimated as $P(h_{k}\mid\vb)\approx\frac{1}{n}\sum_{s=1}^{n}h_{k}^{(s)}$
for $n$ samples.

\paragraph{Variational method.}

We make the approximation 
\[
P(\hb,\ub\mid\vb)\approx\prod_{k}Q_{k}(h_{k})\prod_{i}Q_{i}(u_{i})
\]
Minimising the Kullback-Leibler divergence between $P(\hb,\ub\mid\vb)$
and its approximation leads the following recursive update 
\begin{align}
Q_{k}\left(h_{k}^{(t+1)}=1\right) & \leftarrow\frac{1}{1+\exp\left(-\gamma_{k}-\sum_{i}w_{ik}\left\langle u_{i}\right\rangle _{Q_{i}^{(t)}}\right)}\label{eq:utility-mean-field}\\
Q_{i}\left(u_{i}^{(t+1)}\right) & \leftarrow\frac{1}{\kappa_{i}^{(t)}}\mathbb{I}\left[\u_{i}\in\Omega(v_{i})\right]\mathcal{N}\left(\x_{i};\hat{\mu}{}_{i}(\hb^{(t)}),\sigma_{i}\right)
\end{align}
where $t$ is the update index of the recursion, $\left\langle u_{i}\right\rangle _{Q_{i}^{(t)}}$
is the mean of utility $u_{i}$ with respect to $Q_{i}\left(u_{i}^{(t)}\right)$,
$\kappa_{i}^{(t)}=\int_{u_{i}\in\Omega(v_{i})}\mathcal{N}\left(\x_{i};\mu_{i}(\hb^{(t)}),\sigma_{i}\right)$
is the normalising constant, and $\hat{\mu}{}_{i}(\hb^{(t)})=\sigma_{i}^{2}\left(\alpha_{i}+\sum_{k=1}^{K}w_{ik}Q_{k}\left(h_{k}^{(t)}=1\right)\right)$.
Finally, we obtain $P(h_{k}\mid\vb)\approx Q_{k}\left(h_{k}=1\right)$.

\subsection{Prediction \label{sub:Prediction}}

An important task is \emph{prediction} of the ordinal level of an
unseen variable given the other seen variables, where we need to estimate
the following predictive distribution 
\begin{equation}
P(v_{j}\mid\vb)=\sum_{\hb}\int_{u_{j}\in\Omega(v_{j})}\int_{\ub\in\Domainb(\vb)}P(\hb,u_{j},\ub\mid\vb)d\ub du_{j}\label{eq:prediction}
\end{equation}
Unfortunately, now $(h_{1},h_{2},...,h_{K})$ are coupled due to the
integration over $\{u_{j},\ub\}$ making the evaluation intractable,
and thus approximation is needed.

For simplicity, we assume that the seen data $\vb$ is informative
enough so that $P(\hb|v_{j},\vb)\approx P(\hb|\vb)$. Thus we can
rewrite Eq.~(\ref{eq:prediction}) as

\begin{align*}
P(v_{j}|\vb) & \approx\sum_{\hb}P(\hb|\vb)P(v_{j}|\hb)du_{j}
\end{align*}
Now we make further approximations to deal with the exponential sum
over $\hb$.

\paragraph{MCMC.}

Given the sampling from $P(\hb|\vb)$ described in Section~\ref{sub:Factor-Posteriors},
we obtain

\begin{align*}
P(v_{j}|\vb) & \approx\frac{1}{n}\sum_{s=1}^{n}P(v_{j}|\hb^{(s)})du_{j}
\end{align*}
where $n$ is the sample size, and $P(v_{j}|\hb^{(s)})$ is computed
using Eq.~(\ref{eq:ordinal-gen-given-rep}).

\paragraph{Variational method.}

The idea is similar to mean-field described in Section~\ref{sub:Factor-Posteriors}.
In particular, we estimate $\hat{h}_{k}=P(h_{k}=1|\vb)$ using either
MCMC sampling or mean-field update. The predictive distribution is
approximated as 
\begin{align*}
P(v_{j}|\vb) & \approx\int_{u_{i}\in\Omega(v_{j})}P(u_{i}\mid\hat{h}_{1},\hat{h}_{2},...,\hat{h}_{K})
\end{align*}
where $P(u_{i}\mid\hat{h}_{1},\hat{h}_{2},...,\hat{h}_{K})=\mathcal{N}\left(\x_{i};\sigma_{i}^{2}\left(\alpha_{i}+\sum_{k=1}^{K}w_{ik}\hat{h}_{k}\right),\sigma_{i}\right)$.
The computation is identical to that of Eq.~(\ref{eq:ordinal-gen-given-rep})
if we replace $h_{k}$ (binary) by $\hat{h}_{k}$ (real-valued) .

\subsection{Stochastic Gradient Learning with Persistent Markov Chains \label{sub:Stochastic-Gradient-Learning}}

Learning is based on maximising the data log-likelihood 
\begin{eqnarray*}
\LL & = & \log P(\vb)=\log\sum_{\hb}\int_{\xb}P(\vb,\xb,\hb)d\xb\\
 & = & \log Z(\vb)-\log Z
\end{eqnarray*}
where $P(\vb,\xb,\hb)$ is defined in Eq.~(\ref{eq:Model-def}) and
$Z(\vb)=\sum_{\hb}\int_{\xb\in\Domainb(\vb)}\Psi(\xb,\hb)d\xb$. Note
that $Z(\vb)$ includes $Z$ as a special case when the domain $\Domainb(\vb)$
is the whole real space $\Real^{N}$.

Recall that the model belongs to the exponential family in that we
can rewrite the potential function as

\[
\Psi(\xb,\hb)=\exp\left\{ \sum_{a}W_{a}f_{a}(\xb,\hb)\right\} 
\]
where $f_{a}(\xb,\hb)\in\left\{ u_{i},u_{i}h_{k},h_{k}\right\} _{(i,k)=(1,1)}^{(N,K)}$
is a sufficient statistic, and $W_{a}\in\left\{ \alpha_{i},\gamma_{k},w_{ik}\right\} _{(i,k)=(1,1)}^{(N,K)}$
is its associated parameter. Now the gradient of the log-likelihood
has the standard form of difference of expected sufficient statistics
(ESS) 
\begin{eqnarray*}
\partial_{W_{a}}\LL & = & \left\langle f_{a}\right\rangle _{P(\xb,\hb\mid\vb)}-\left\langle f_{a}\right\rangle _{P(\xb,\hb)}
\end{eqnarray*}
where $P(\xb,\hb\mid\vb)$ is a truncated Gaussian RBM and $P(\xb,\hb)$
is the standard Gaussian RBM.

Put in common RBM-terms, there are two learning phases: the \emph{clamped
phase} in which we estimate the ESS w.r.t. the empirical distribution
$P\left(\xb,\hb\mid\vb\right)$, and the \emph{free phase} in which
we compute the ESS w.r.t. model distribution $P(\xb,\hb)$.

\subsubsection{Persistent Markov Chains}

The literature offers efficient stochastic gradient procedures to
learn parameters, in which the method of \cite{younes1989parametric}
and its variants -- the Contrastive Divergence of \cite{Hinton02}
and its persistent version of \cite{tieleman2008training} -- are
highly effective in large-scale settings. The strategy is to update
parameters after short Markov chains. Typically only the free phase
requires the MCMC approximation. In our setting, on the other hand,
both the clamped phase and the free phase require approximation.

Since it is possible to integrate over utilities when the binary factors
are known, it is tempting so sample only the binary factors in the
Rao-Blackwellisation fashion. However, here we take the advantage
of the bipartite structure of the underlying RBM: the layer-wise sampling
is efficient and much simpler. Once the hidden factor samples are
obtained, we integrate over utilities for better numerical stability.
The ESSes are the averaged over all factor samples.

For the clamped phase, we maintain one Markov chain per data instance.
For memory efficiency, only the binary factor samples are stored between
update steps. For the free phase, there are two strategies: 
\begin{itemize}
\item \emph{Contrastive} \emph{chains}: one short chain is needed per data
instance, but initialised from the clamped chain. That is, we discard
those chains after each update. 
\item \emph{Persistent chains}: free-phase chains are maintained during
the course of learning, independent of the clamp-phase chains. If
every data instance has the same dimensions (which they do not, in
the case of missing data), we need to maintain a moderate number of
chains (e.g., $20-100$). Otherwise, we need one chain per data instance. 
\end{itemize}
At each step, we collect a small number of samples and estimate the
approximate distributions $\tilde{P}(\xb,\hb\mid\vb)$ and $\tilde{P}(\xb,\hb)$.
The parameters are updated according to the stochastic gradient ascent
rule 
\[
W_{s}\leftarrow W_{s}+\nu\left(\left\langle f_{a}\right\rangle _{\tilde{P}(\xb,\hb|\vb)}-\left\langle f_{a}\right\rangle _{\tilde{P}(\xb,\hb)}\right)
\]
where $\nu\in(0,1)$ is the learning rate.

\subsubsection{Learning Thresholds}

Thresholds appear only in the computation of $Z(\vb)$ as they define
the utility domain $\Domainb(\vb)$. Let $\bar{\Domainb}(\vb)^{+}$
be the upper boundary of $\Domainb(\vb)$, and $\bar{\Domainb}(\vb)^{-}$
the lower boundary. The gradient of the log-likelihood w.r.t. boundaries
reads 
\begin{eqnarray*}
\partial_{\bar{\Domainb}(\vb)^{+}}\LL & = & \frac{1}{Z(\vb)}\sum_{\hb}\partial_{\bar{\Domainb}(\vb)^{+}}\int_{\xb\in\Domainb(\vb)}\Psi(\hb,\xb)d\ub=\sum_{\hb}P(\xb=\bar{\Domainb}(\vb)^{+},\hb\mid\vb)\\
\partial_{\bar{\Domainb}(\vb)^{-}}\LL & = & -\sum_{\hb}P(\xb=\bar{\Domainb}(\vb)^{-},\hb\mid\vb)
\end{eqnarray*}
Recall from Section~\ref{sub:Model-Structure-and-Param} that the
boundaries $\bar{\Omega}(v_{i}=l)^{-}$ and $\bar{\Omega}(v_{i}=l)^{+}$
are the lower-threshold $\theta_{i(l-1)}$ and the upper-threshold
$\theta_{il}$, respectively, where $\theta_{il}=\theta_{i(l-1)}+e^{\tau_{il}}=\tau_{i1}+\sum_{m=2}^{l}e^{\tau_{im}}$.
Using the chain rule, we would derive the derivatives w.r.t. to $\left\{ \tau_{im}\right\} _{m=2}^{L-1}$.

\subsection{Handling Heterogeneous Data \label{sub:Handling-Heterogeneous-Data}}

We now consider the case where ordinal variables do not share the
same ordinal scales, that is, we have a separate ordered set $S_{i}=\{c_{i1}\prec c_{i2}\prec...,\prec c_{iL_{i}}\}$
for each variable $i$. This requires only slight change from the
homogeneous case, e.g., by learning separate set of thresholds for
each variable.

\section{$\Model$ for Matrix Data \label{sec:CBM-for-Matrix}}

Often the data has the matrix form, i.e., a list of column vectors
and we often assume columns as independent. However, this assumption
is too strong in many applications. For example, in collaborative
filtering where each user plays the role of a column, and each item
the role of a row, a user's choice can be influenced by other users'
choices (e.g., due to the popularity of a particular item), then columns
are correlated. Second, it is also natural to switch the roles of
the users and items and this clearly destroys the i.i.d assumption
over the columns.

Thus, it is more precise to assume that an observation is \emph{jointly}
generated by both the row-wise and column-wise processes \cite{Truyen:2009a}.
In particular, let $d$ be the index of the data instance, each observation
$v_{di}$ is generated from an utility $u_{di}$. Each data instance
(column) $d$ is represented by a vector of binary hidden factors
$\hb_{d}\in\left\{ 0,1\right\} ^{K}$ and each item (row) $i$ is
represented by a vector of binary hidden factors $\gb_{i}\in\left\{ 0,1\right\} ^{S}$.
Since our data matrix is usually incomplete, let us denote by $W\in\left\{ 0,1\right\} ^{D\times N}$
the incidence matrix where $W_{di}=1$ if the cell $(d,i)$ is observed,
and $W_{di}=0$ otherwise. There is a single model for the whole incomplete
data matrix. Every observed entry $(d,i)$ is connected with two sets
of hidden factors $\hb_{d}$ and $\gb_{i}$. Consequently, there are
$DK+NS$ binary factor units in the entire model.

Let $\Hb=\left(\left\{ u_{di}\right\} _{W_{di}=1},\left\{ \hb_{d}\right\} _{d=1}^{D},\left\{ \gb_{i}\right\} _{i=1}^{N}\right)$
denote all latent variables and $\Vb=\left\{ v_{di}\right\} _{W_{di}=1}$
all visible ordinal variables. The matrix-variate model distribution
has the usual form 
\[
P(\Vb,\Hb)=\frac{1}{Z^{*}}\Psi^{*}\left(\Hb\right)\prod_{d,i|W_{di}=1}P(v_{di}\mid u_{di})
\]
where $Z^{*}$ is the normalising constant and $\Psi^{*}\left(\Hb\right)$
is the product of all local potentials. More specifically, 
\begin{eqnarray*}
\Psi^{*}\left(\Hb\right) & = & \prod_{d,i|W_{di}=1}\left(\phi_{di}(\x_{di})\prod_{k}\psi_{ik}(\x_{di},h_{dk})\prod_{s}\varphi_{is}(\x_{di},g_{is})\right)\left[\prod_{d,k}\phi_{k}(h_{dk})\right]\left[\prod_{i,s}\phi_{s}(g_{is})\right]
\end{eqnarray*}
where $\psi_{ik}(\x_{di},h_{k}),\phi_{k}(h_{dk})$ are the same as
those defined in Eq.~(\ref{eq:poten-params}), respectively, and
{\small 
\begin{eqnarray*}
\phi_{di}(\x_{di}) & = & \exp\left\{ -\frac{\x_{di}^{2}}{2\sigma_{di}^{2}}+(\alpha_{i}+\beta_{d})\x_{i}\right\} ;\quad\varphi_{ds}(\x_{di},g_{s})=\exp\left\{ \omega_{ds}u_{di}g_{s}\right\} ;\quad\phi_{s}(g_{is})=\exp\left\{ \xi_{s}g_{is}\right\} 
\end{eqnarray*}
}{\small \par}

The ordinal model $P(v_{di}\mid u_{di})$ is similar to that defined
in Eq.~(\ref{eq:ordinal-model}) except for the thresholds, which
are now functions of both the data instance and the item, that is
$\theta_{di1}=\tau_{i1}+\kappa_{d1}$ and $\theta_{dil}=\theta_{di(l-1)}+e^{\tau_{il}+\kappa_{dl}}$
for $l=2,3,..,L-1$.

\subsection{Model Properties}

It is easy to see that conditioned on the utilities, the posteriors
of the binary factors are still factorisable. Likewise, given the
factors, the utilities are univariate Gaussian

\begin{eqnarray*}
P(u_{di}\mid\hb_{d},\gb_{i}) & = & \Normal\left(\mu_{di}^{*}(\hb_{d},\gb_{i}),\sigma_{di}^{2}\right)\\
P(u_{di}\mid\hb_{d},\gb_{i},v_{di}) & \propto & \mathbb{I}\left[\u_{di}\in\Omega(v_{di})\right]P(u_{di}\mid\hb_{d},\gb_{i})
\end{eqnarray*}
where $\Omega(v_{di})$ is the domain defined by the thresholds at
the level $l=v_{di}$, and the mean structure is 
\begin{equation}
\mu_{di}^{*}(\hb_{d},\gb_{i})=\sigma_{di}^{2}\left(\alpha_{i}+\beta_{d}+\sum_{k=1}^{K}w_{ik}h_{dk}+\sum_{s=1}^{S}\omega_{ds}g_{is}\right)\label{eq:mean-structure-matrix}
\end{equation}

Previous inference tricks can be re-used by noting that for each column
(i.e., data instance), we still enjoy the Gaussian RBM when conditioned
on other columns\emph{.} The same holds for rows\emph{ }(i.e., items)\emph{.}

\subsection{Stochastic Learning with Structured Mean-Fields}

Although it is possible to explore the space of the whole model using
Gibbs sampling and use the short MCMC chains as before, here we resort
to structured mean-field methods to exploit the modularity in the
model structure. The general idea is to alternate between the column-wise
and the row-wise conditional processes: 
\begin{itemize}
\item In the \emph{column-wise} process, we estimate item-specific factor
posteriors $\left\{ \hat{\gb_{i}}\right\} _{i=1}^{N}$, where $\hat{g}_{is}\leftarrow P\left(g_{is}=1\mid\left(v_{di}\right)_{di\mid W_{id}=1}\right)$
and use them \emph{as if} the item-specific factors $\left(\gb_{i}\right)_{i=1}^{N}$
are given. For example, the mean structure in Eq.~(\ref{eq:mean-structure-matrix})
now has the following form{\small 
\[
\mu_{di}^{*}(\hb_{d},\hat{\gb}_{i})=\sigma_{di}^{2}\left(\alpha_{i}+\beta_{d}+\sum_{k=1}^{K}w_{ik}h_{dk}+\sum_{s=1}^{S}\omega_{ds}\hat{g}_{is}\right)
\]
}which is essentially the mean structure in Eq.~(\ref{eq:mean-structure})
when $\beta_{d}+\sum_{s=1}^{S}\omega_{ds}\hat{g}_{is}$ is absorbed
into $\alpha_{i}$. Conditioned on the estimated posteriors, the data
likelihood is now factorisable $\prod_{d}P\left(\vb_{d\bullet}\mid\left\{ \hat{\gb_{i}}\right\} _{i=1}^{N}\right)$,
where $\vb_{d\bullet}$ denotes the observations of the $d$-th data
instance. 
\item Similarly, in the \emph{row-wise} process we estimate data-specific
posteriors $\left\{ \hat{\hb_{d}}\right\} _{d=1}^{D}$, where $\hat{h}_{dk}=P\left(h_{dk}=1\mid\left(v_{di}\right)_{W_{id}=1}\right)$
and use them \emph{as if} the data-specific factors $\left(\hb_{d}\right)_{d=1}^{D}$
are given. The data likelihood has the form $\prod_{i}P\left(\vb_{\bullet i}\mid\left\{ \hat{\hb_{d}}\right\} _{d=1}^{D}\right)$,
where $\vb_{\bullet i}$ denotes the observations of the $i$-th item. 
\end{itemize}
At each step, we then improve the conditional data likelihood using
the gradient technique described in Section~\ref{sub:Stochastic-Gradient-Learning},
e.g., by running through the whole data once.

\subsubsection{Online Estimation of Posteriors \label{sub:Online-Estimation-of-Posteriors}}

The structured mean-fields technique requires the estimation of the
factor posteriors. To reduce computation, we propose to treat the
trajectory of the factor posteriors during learning as a \emph{stochastic
process}. This suggests a simple smoothing method, e.g., at step $t$:

\[
\hat{\hb}_{d}^{(t)}\leftarrow\eta\hat{\hb}_{d}^{(t-1)}+(1-\eta)P\left(h_{dk}=1\mid\ub_{d}^{(t)}\right)
\]
where $\eta\in(0,1)$ is the \emph{smoothing factor, }and $\ub_{d}^{(t)}$
is a utility sample in the \emph{clamped phase}. This effectively
imposes an exponential decay to previous samples. The estimation of
$\eta$ would be of interest in its own right, but we would empirically
set $\eta\in(0.5,0.9)$ and do not pursue the issue further.

\section{Experiments \label{sec:Experiments}}

In this section, we demonstrate how $\Model$ can be useful in real-world
data analysis tasks. To monitor learning progress, we estimate the
data pseudo-likelihood $P(v_{i}\mid\vb_{\neg i})$. For simplicity,
we treat $v_{i}$ as if it is not in $\vb$ and replace $\vb_{i}$
by $\vb$. This enables us to use the same predictive methods in Section~\ref{sub:Prediction}.
See Fig.~\ref{fig:Vector-versus-matrix}(a) for an example of the
learning curves. To sample from the truncated Gaussian, we employ
methods described in \cite{robert1995simulation}, which is more efficient
than standard rejection sampling techniques. Mapping parameters $\left\{ w_{ik}\right\} $
are initialised randomly, bias paramters are from zeros, and thresholds
$\left\{ \theta_{il}\right\} $ are spaced evenly at the begining.

\subsection{Global Attitude Analysis: Latent Profile Discovery}

\begin{figure}
\begin{centering}
\includegraphics[width=0.6\textwidth,height=0.45\textwidth]{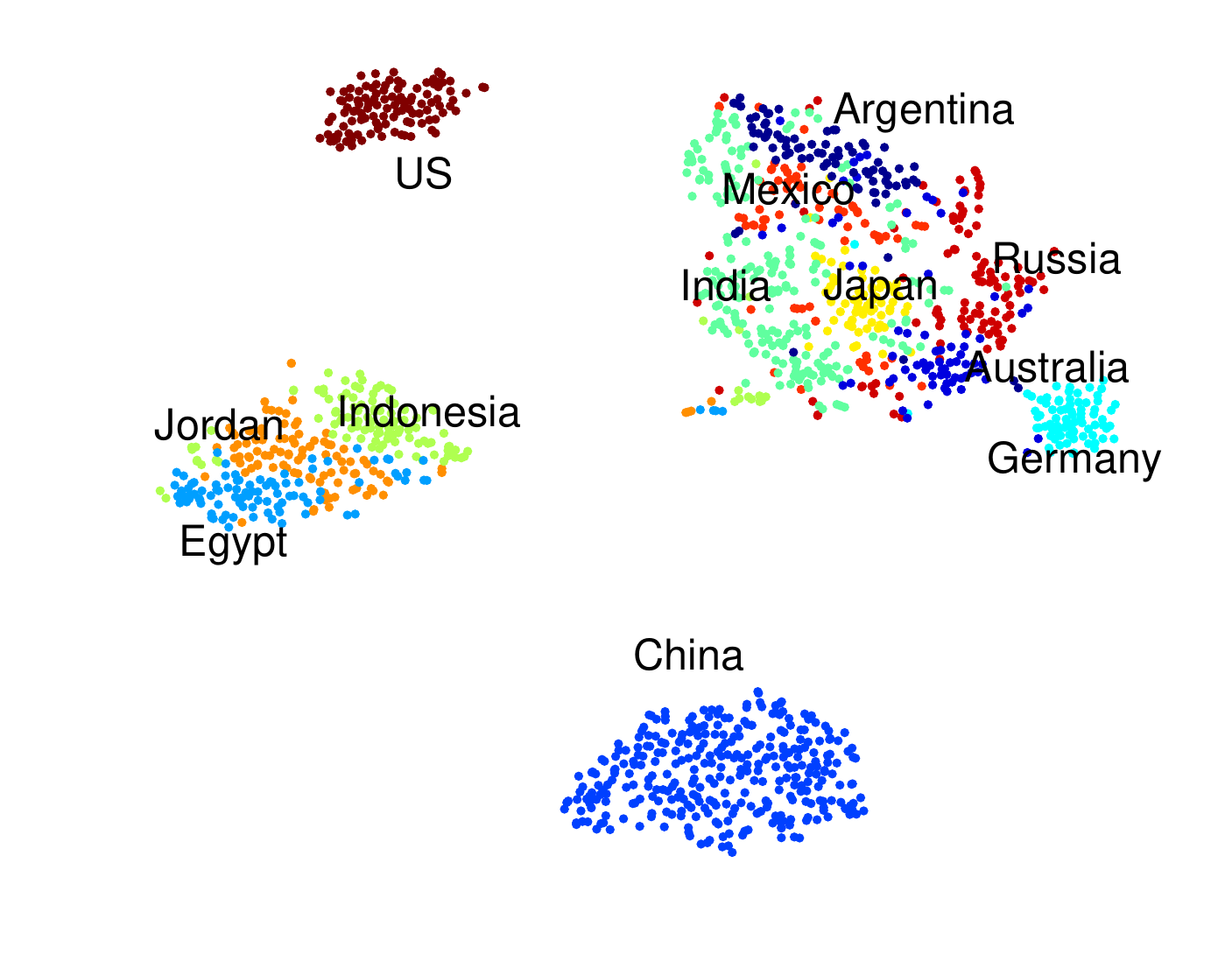} 
\par\end{centering}

\caption{Visualisation of world's opinions in 2008 by projecting latent posteriors
$\hat{\hb}=\left(P(h_{1}^{1}|\vb),P(h_{2}^{1}|\vb,...,P(h_{K}^{1}|\vb)\right)$
on 2D using t-SNE \cite{van2008visualizing}, where $h_{k}^{1}$ is
a shorthand for $h_{k}=1$. Best viewed in colours. \label{fig:Visualisation-of-world-opinions}.}
\end{figure}

In this experiments we validate the capacity to discover meaningful
latent profiles from people's opinions about their life and the social/political
conditions in their country and around the world. We use the public
world-wide survey by PewResearch Centre%
\footnote{http://pewresearch.org/%
} in 2008 which interviewed $24,717$ people from $24$ countries.
After re-processing, we keep $165$ ordinal responses per respondent.
Example questions are: ``(Q1) {[}..{]} how would you describe your
day today\textemdash{}has it been a typical day, a particularly good
day, or a particularly bad day?'', ``(Q5) {[}...{]} over the next
12 months do you expect the economic situation in our country to improve
a lot, improve a little, remain the same, worsen a little or worsen
a lot?''.

The data is heterogeneous since question types are different (see
Section~\ref{sub:Handling-Heterogeneous-Data}). For this we use
a vector-based $\Model$ with $K=50$ hidden units. After model fitting,
we obtain a posterior vector $\hat{\hb}=\left(P(h_{1}^{1}|\vb),P(h_{2}^{1}|\vb,...,P(h_{K}^{1}|\vb)\right)$,
which is then used as the representation of the respondent's latent
profile. For visualisation, we project this vector onto the 2D plane
using a locality-preserving dimensionality reduction method known
as \emph{t-SNE}%
\footnote{Note that the t-SNE does not do clustering, it tries only to map from
the input to the 2D so that local properties of the data in preserved.%
} \cite{van2008visualizing}. The opinions of citizens of $12$ countries
are depicted in Fig.~\ref{fig:Visualisation-of-world-opinions}.
This clearly reveals how cultures (e.g., Islamic and Chinese) and
nations (e.g., the US, China, Latin America) see the world.

\subsection{Collaborative Filtering: Matrix Completion}

We verify our models on three public rating datasets: MovieLens%
\footnote{http://www.grouplens.org/node/12%
} -- containing $1$ million ratings by $6$ thousand users on nearly
$4$ thousand movies; Dating%
\footnote{http://www.occamslab.com/petricek/data/%
} -- consisting of $17$ million ratings by $135$ thousand users on
nearly $169$ thousand profiles; and Netflix%
\footnote{http://netflixprize.com/%
} -- $100$ millions ratings by $480$ thousand users on nearly $18$
thousand movies. The Dating ratings are on the $10$-point scale and
the other two are on the $5$-star scale. We then transform the Dating
ratings to the $5$-point scale for uniformity. For each data we remove
those users with less than $30$ ratings, $5$ of which are used for
tuning and stopping criterion, $10$ for testing and the rest for
training. For MovieLens and Netflix, we ensure that rating timestamps
are ordered from training, to validation to testing. For the Dating
dataset, the selection is at random.

For comparison, we implement state-of-the-art methods in the field,
including: Matrix Factorisation (MF) with Gaussian assumption \cite{salakhutdinov2008probabilistic},
MF with cumulative ordinal assumption \cite{koren2011ordrec} (without
item-item neighbourhood), and RBM with multinomial assumption \cite{Salakhutdinov-et-alICML07}.For
prediction in the CRBM, we employ the variational method (Section~\ref{eq:prediction}).
The training and testing protocols are the same for all methods: Training
stops where there is no improvement on the likelihood of the validation
data. Two popular performance metrics are reported on the test data:\emph{
}the \emph{root-mean square error} (RMSE), the \emph{mean absolute
error} (MAE). Prediction for ordinal MF and RBMs is a numerical mean
in the case of RMSE: $v_{j}^{RMSE}=\sum_{l=1}^{L}P(v_{j}=l|\vb)l$,
and an MAP estimation in the case of MAE: $v_{j}^{MAE}=\arg\max_{l}P(v_{j}=l|\vb)$.

\begin{figure}
\begin{centering}
\begin{tabular}{ccc}
\includegraphics[width=0.4\textwidth,height=0.35\textwidth]{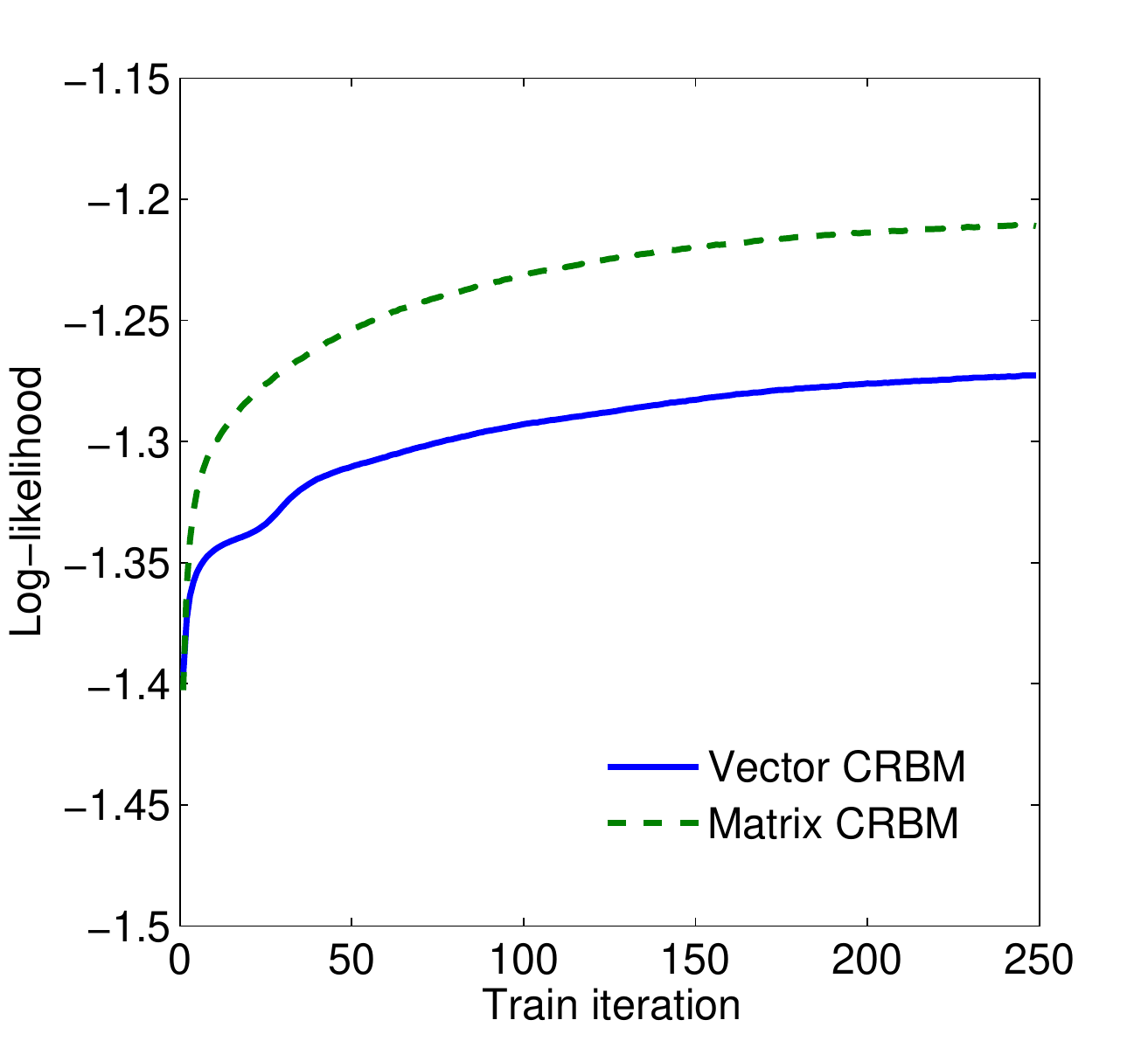}  &  & \includegraphics[width=0.4\textwidth,height=0.35\textwidth]{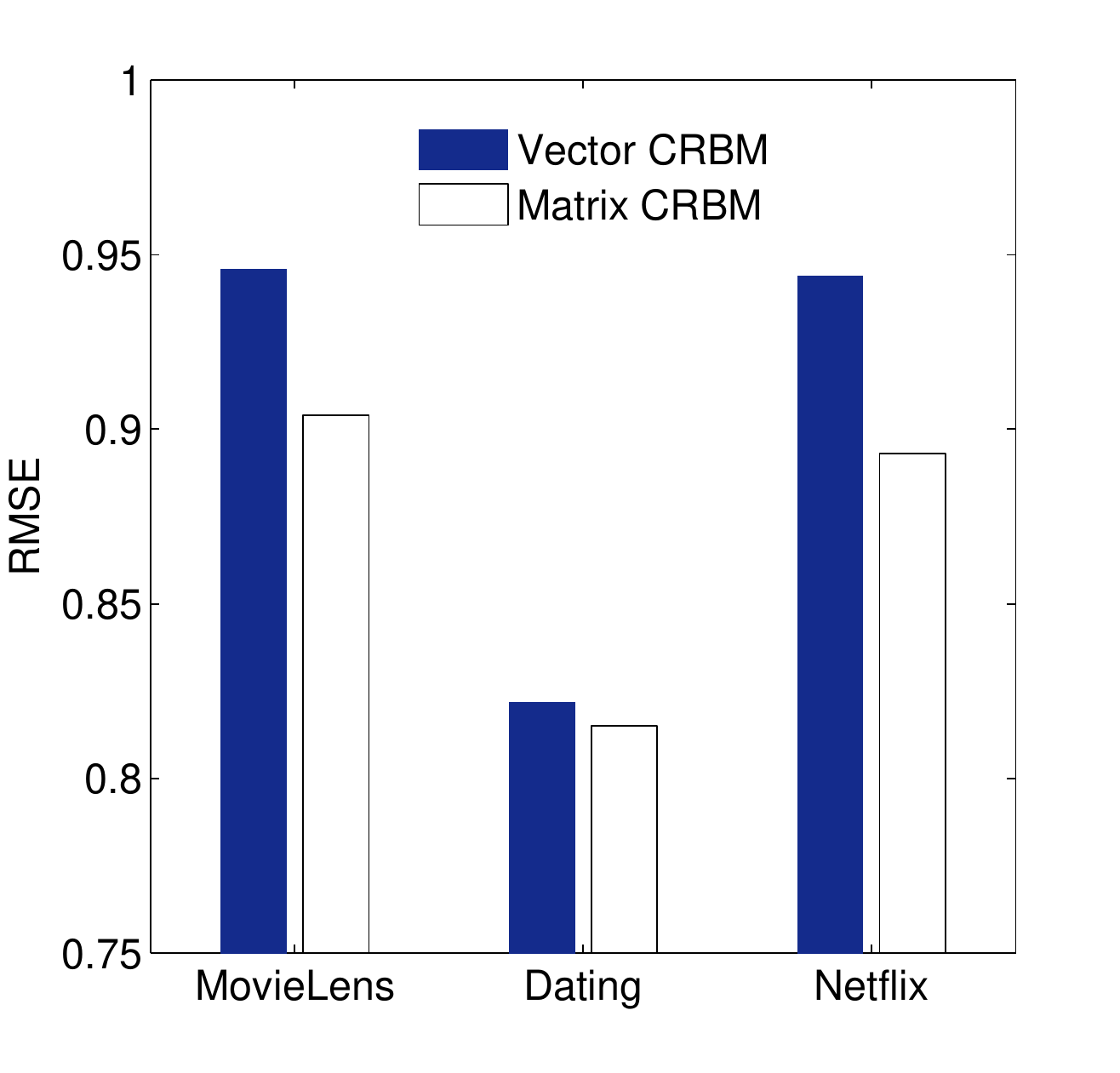}\tabularnewline
(a) Monitoring pseudo-likelihood in training  &  & (b) RMSE on test data\tabularnewline
\end{tabular}
\par\end{centering}

\caption{Vector versus matrix $\Model$s, where $K=50$.\label{fig:Vector-versus-matrix}}
\end{figure}

Fig.~\ref{fig:Vector-versus-matrix}(a) depicts the learning curve
of the vector-based and matrix-based $\Model$s, and Fig.~\ref{fig:Vector-versus-matrix}(b)
shows their predictive performance on test datasets. Clearly, the
effect of matrix treatment is significant. Tables~\ref{tab:MovieLens-1M},\ref{tab:Dating-16M},\ref{tab:Netflix-100M}
report the performances of all methods on the three datasets. The
(matrix) $\Model$ are often comparable with the best rivals on the
RMSE scores and are competitive against all others on the MAE.

\begin{table*}
\begin{centering}
\begin{tabular}{|c|c|c|c|c|c|c|}
\hline 
\textcolor{black}{{} }  & \multicolumn{2}{c|}{\textcolor{black}{$K=50$}} & \multicolumn{2}{c|}{\textcolor{black}{$K=100$}} & \multicolumn{2}{c|}{\textcolor{black}{$K=200$}}\tabularnewline
\hline 
 & \textcolor{black}{\emph{RMSE}}  & \textcolor{black}{\emph{MAE}}  & \textcolor{black}{\emph{RMSE}}  & \textcolor{black}{\emph{MAE}}  & \textcolor{black}{\emph{RMSE}}\textcolor{black}{{} }  & \textcolor{black}{\emph{MAE}}\tabularnewline
\hline 
\hline 
\textcolor{black}{Gaussian Matrix Fac.}  & \textcolor{black}{0.914 }  & \textcolor{black}{0.720 }  & \textcolor{black}{0.911 }  & \textcolor{black}{0.719 }  & \textcolor{black}{0.908}  & \textcolor{black}{0.716}\tabularnewline
\textcolor{black}{Ordinal Matrix Fac.}  & \textbf{\textcolor{black}{0.904}}  & \textcolor{black}{0.682}  & \textbf{\textcolor{black}{0.902}}  & \textcolor{black}{0.682}  & \textbf{\textcolor{black}{0.902}}\textcolor{black}{{} }  & \textcolor{black}{0.680}\tabularnewline
\textcolor{black}{Multinomial RBM}  & \textcolor{black}{0.928}  & \textcolor{black}{0.711}  & \textcolor{black}{0.926}  & \textcolor{black}{0.707}  & \textcolor{black}{0.928}  & \textcolor{black}{0.708}\tabularnewline
\hline 
\textcolor{black}{Matrix Cumul. RBM }  & \textbf{\textcolor{black}{0.904}}  & \textbf{\textcolor{black}{0.666}}  & \textcolor{black}{0.904}  & \textbf{\textcolor{black}{0.662}}\textcolor{black}{{} }  & \textcolor{black}{0.906 }  & \textbf{\textcolor{black}{0.664}}\tabularnewline
\hline 
\end{tabular}
\par\end{centering}

\caption{Results on MovieLens (the smaller the better). \label{tab:MovieLens-1M}}
\end{table*}

\begin{table*}
\begin{centering}
\begin{tabular}{|c|c|c|c|c|c|c|}
\hline 
\textcolor{black}{{} }  & \multicolumn{2}{c|}{\textcolor{black}{$K=50$}} & \multicolumn{2}{c|}{\textcolor{black}{$K=100$}} & \multicolumn{2}{c|}{\textcolor{black}{$K=200$}}\tabularnewline
\hline 
 & \textcolor{black}{\emph{RMSE}}  & \textcolor{black}{\emph{MAE}}  & \textcolor{black}{\emph{RMSE}}  & \textcolor{black}{\emph{MAE}}  & \textcolor{black}{\emph{RMSE}}\textcolor{black}{{} }  & \textcolor{black}{\emph{MAE}}\tabularnewline
\hline 
\hline 
\textcolor{black}{Gaussian Matrix Fac. }  & \textcolor{black}{0.852 }  & \textcolor{black}{0.596 }  & \textcolor{black}{0.848 }  & \textcolor{black}{0.592 }  & \textcolor{black}{0.840 }  & \textcolor{black}{0.586}\tabularnewline
\textcolor{black}{Ordinal Matrix Fac.}  & \textcolor{black}{0.857 }  & \textcolor{black}{0.511 }  & \textcolor{black}{0.854 }  & \textcolor{black}{0.507 }  & \textcolor{black}{0.849 }  & \textcolor{black}{0.502}\tabularnewline
\textcolor{black}{Multinomial RBM}  & \textbf{\textcolor{black}{0.815}}  & \textcolor{black}{0.483}  & \textbf{\textcolor{black}{0.794}}  & \textcolor{black}{0.470}  & 0.787  & 0.463\tabularnewline
\hline 
\textcolor{black}{Matrix Cumul. RBM }  & \textbf{\textcolor{black}{0.815}}  & \textbf{\textcolor{black}{0.475}}  & \textcolor{black}{0.799}  & \textbf{\textcolor{black}{0.461}}\textcolor{black}{{} }  & \textcolor{black}{0.794 }  & \textbf{\textcolor{black}{0.458}}\tabularnewline
\hline 
\end{tabular}
\par\end{centering}

\caption{Results on Dating (the smaller the better). \label{tab:Dating-16M}}
\end{table*}

\begin{table}
\begin{centering}
\begin{tabular}{|c|c|c|c|c|}
\hline 
 & \multicolumn{2}{c|}{\textcolor{black}{$K=50$}} & \multicolumn{2}{c|}{\textcolor{black}{$K=100$}}\tabularnewline
\hline 
 & \textcolor{black}{\emph{RMSE}}  & \textcolor{black}{\emph{MAE}}  & \textcolor{black}{\emph{RMSE}}\textcolor{black}{{} }  & \textcolor{black}{\emph{MAE}}\tabularnewline
\hline 
\hline 
\textcolor{black}{Gaussian Matrix Fac. }  & \textbf{\textcolor{black}{0.890}}  & \textcolor{black}{0.689}  & \textcolor{black}{0.888}\textbf{\textcolor{black}{{} }}  & \textcolor{black}{0.688}\tabularnewline
\textcolor{black}{Ordinal Matrix Fac. }  & \textcolor{black}{0.904}  & \textcolor{black}{0.658 }  & \textcolor{black}{0.902 }  & \textcolor{black}{0.657}\tabularnewline
\textcolor{black}{Multinomial RBM}  & \textcolor{black}{0.894}  & \textcolor{black}{0.659}  & \textbf{0.887}  & 0.650\tabularnewline
\hline 
\textcolor{black}{Matrix Cumul. RBM }  & \textcolor{black}{0.893}  & \textbf{\textcolor{black}{0.641}}\textcolor{black}{{} }  & \textcolor{black}{0.892 }  & \textbf{\textcolor{black}{0.640}}\tabularnewline
\hline 
\end{tabular}
\par\end{centering}

\caption{Results on Netflix (the smaller the better). \label{tab:Netflix-100M}}
\end{table}

\section{Related Work \label{sec:Related-Work}}

This work partly belongs to the thread of research that extends RBMs
for a variety of data types, including \emph{categories }\cite{Salakhutdinov-et-alICML07}\emph{,
counts }\cite{gehler2006rap,salakhutdinov2009semantic,salakhutdinov2009replicated},
\emph{bounded} variables \cite{le2011learning} and a mixture of these
types \cite{Truyen:2011b}. Gaussian RBMs have been only used for
continuous variables \cite{Hinton02,mohamed2010phone} -- thus our
use for ordinal variables is novel. There has also been recent work
extending Gaussian RBMs to better model highly correlated input variables
\cite{ranzato2010modeling,courville2011spike}. For ordinal data,
to the best of our knowledge, the first RBM-based work is \cite{Truyen:2009a},
which also contains a treatment of matrix-wise data\emph{.} However,
their work indeed models multinomial data with knowledge of orders
rather than modelling the ordinal nature directly. The result is that
it is over-parameterised but less efficient and does not offer any
underlying generative mechanism for ordinal data.

Ordinal data has been well investigated in statistical sciences, especially
quantitative social studies, often under the name of \emph{ordinal
regression}, which refers to single ordinal output given a set of
input covariates. The most popular method is by \cite{mccullagh1980rmo}
which examines the level-wise cumulative distributions. Another well-known
treatment is the sequential approach, also known as continuation ratio
\cite{mare1980social}, in which the ordinal generation process is
considered stepwise, starting from the lowest level until the best
level is chosen. For reviews of recent development, we refer to \cite{liu2005aoc}.
In machine learning, this has attracted a moderate attention in the
past decade \cite{herbrich1999large,chu2006gpo,chu2007svo,cardoso2007learning},
adding machine learning flavours (e.g., large-margins) to existing
statistical methods.

Multivariate ordinal variables have also been studied for several
decades \cite{anderson1985grouped}. The most common theme is the
assumption of the latent multivariate normal distribution that generates
the ordinal observations, often referred to as \emph{multivariate
probit} models \cite{chib1998analysis,Grilli2003,kottas2005nonparametric,podani2005multivariate,jeliazkov2008fitting,chagneau2010hierarchical}.
The main problem with this setting is that it is only feasible for
problems with small dimensions. Our treatment using RBMs offer a solution
for large-scale settings by transferring the low-order interactions
among the Gaussian variables onto higher-order interactions through
the hidden binary layer. Not only this offers much faster inference,
it also enables automatic discovery of latent aspects in the data.

For matrix data, the most well-known method is perhaps matrix factorisation
\cite{lee1999lpo,rennie2005fmm,salakhutdinov2008probabilistic}. However,
this method assumes that the data is normally distributed, which does
not meet the ordinal characteristics well. Recent research has attempted
to address this issue \cite{paquet2011hierarchical,koren2011ordrec,Truyen:2012b}.
In particular, \cite{paquet2011hierarchical,koren2011ordrec} adapt
cumulative models of \cite{mccullagh1980rmo}, and \cite{Truyen:2012b}
tailors the sequential models of \cite{mare1980social} for task.

\section{Conclusion \label{sec:Conclusion}}

We have presented $\Model$, a novel probabilistic model to handle
vector-variate and matrix-variate ordinal data. The model is based
on Gaussian restricted Boltzmann machines and we present the model
architecture, learning and inference procedures. We show that the
model is useful in profiling opinions of people across cultures and
nations. The model is also competitive against state-of-art methods
in collaborative filtering using large-scale public datasets. Thus
our work enriches the RBMs, and extends their use on multivariate
ordinal data in diverse applications.


\begin{thebibliography}{10}

\bibitem{anderson1985grouped}
J.A. Anderson and JD~Pemberton.
\newblock The grouped continuous model for multivariate ordered categorical
  variables and covariate adjustment.
\newblock {\em Biometrics}, pages 875--885, 1985.

\bibitem{marlin2010inductive}
Bo~Chen Benjamin~Marlin, Kevin~Swersky and Nando de~Freitas.
\newblock {Inductive Principles for Restricted Boltzmann Machine Learning}.
\newblock In {\em Proceedings of the 13rd International Conference on
  Artificial Intelligence and Statistics}, Chia Laguna Resort, Sardinia, Italy,
  May 2010.

\bibitem{cardoso2007learning}
J.S. Cardoso and J.F.P. da~Costa.
\newblock Learning to classify ordinal data: the data replication method.
\newblock {\em Journal of Machine Learning Research}, 8(1393-1429):6, 2007.

\bibitem{chagneau2010hierarchical}
P.~Chagneau, F.~Mortier, N.~Picard, and J.N. Bacro.
\newblock A hierarchical bayesian model for spatial prediction of multivariate
  non-gaussian random fields.
\newblock {\em Biometrics}, 2010.

\bibitem{chib1998analysis}
S.~Chib and E.~Greenberg.
\newblock Analysis of multivariate probit models.
\newblock {\em Biometrika}, 85(2):347--361, 1998.

\bibitem{chu2006gpo}
W.~Chu and Z.~Ghahramani.
\newblock {Gaussian processes for ordinal regression}.
\newblock {\em Journal of Machine Learning Research}, 6(1):1019, 2006.

\bibitem{chu2007svo}
W.~Chu and S.S. Keerthi.
\newblock {Support vector ordinal regression}.
\newblock {\em Neural computation}, 19(3):792--815, 2007.

\bibitem{courville2011spike}
A.~Courville, J.~Bergstra, and Y.~Bengio.
\newblock {A spike and slab restricted Boltzmann machine}.
\newblock In {\em Proceedings of the 14th International Conference on
  Artificial Intelligence and Statistics (AISTATS)}, volume~15. Fort
  Lauderdale, USA, 2011.

\bibitem{freund1994unsupervised}
Y.~Freund and D.~Haussler.
\newblock Unsupervised learning of distributions on binary vectors using two
  layer networks.
\newblock {\em Advances in Neural Information Processing Systems}, pages
  912--919, 1993.

\bibitem{gehler2006rap}
P.V. Gehler, A.D. Holub, and M.~Welling.
\newblock {The rate adapting Poisson model for information retrieval and object
  recognition}.
\newblock In {\em Proceedings of the 23rd international conference on Machine
  learning}, pages 337--344. ACM New York, NY, USA, 2006.

\bibitem{Grilli2003}
Leonardo Grilli and Carla Rampichini.
\newblock Alternative specifications of multivariate multilevel probit ordinal
  response models.
\newblock {\em Journal of Educational and Behavioral Statistics}, 28(1):31--44,
  2003.

\bibitem{herbrich1999large}
R.~Herbrich, T.~Graepel, and K.~Obermayer.
\newblock {Large margin rank boundaries for ordinal regression}.
\newblock {\em Advances in Neural Information Processing Systems}, pages
  115--132, 1999.

\bibitem{Hinton02}
G.E. Hinton.
\newblock Training products of experts by minimizing contrastive divergence.
\newblock {\em Neural Computation}, 14:1771--1800, 2002.

\bibitem{hinton2006rdd}
G.E. Hinton and R.R. Salakhutdinov.
\newblock Reducing the dimensionality of data with neural networks.
\newblock {\em Science}, 313(5786):504--507, 2006.

\bibitem{jeliazkov2008fitting}
I.~Jeliazkov, J.~Graves, and M.~Kutzbach.
\newblock Fitting and comparison of models for multivariate ordinal outcomes.
\newblock {\em Advances in Econometrics}, 23:115--156, 2008.

\bibitem{koren2011ordrec}
Y.~Koren and J.~Sill.
\newblock {OrdRec: an ordinal model for predicting personalized item rating
  distributions}.
\newblock In {\em Proceedings of the fifth ACM conference on Recommender
  systems}, pages 117--124. ACM, 2011.

\bibitem{kottas2005nonparametric}
A.~Kottas, P.~M{\"u}ller, and F.~Quintana.
\newblock {Nonparametric Bayesian modeling for multivariate ordinal data}.
\newblock {\em Journal of Computational and Graphical Statistics},
  14(3):610--625, 2005.

\bibitem{larochelle2008classification}
H.~Larochelle and Y.~Bengio.
\newblock {Classification using discriminative restricted Boltzmann machines}.
\newblock In {\em Proceedings of the 25th international conference on Machine
  learning}, pages 536--543. ACM, 2008.

\bibitem{le2011learning}
Q.V. Le, W.Y. Zou, S.Y. Yeung, and A.Y. Ng.
\newblock Learning hierarchical invariant spatio-temporal features for action
  recognition with independent subspace analysis.
\newblock {\em CVPR}, 2011.

\bibitem{le2008representational}
N.~Le~Roux and Y.~Bengio.
\newblock {Representational power of restricted Boltzmann machines and deep
  belief networks}.
\newblock {\em Neural Computation}, 20(6):1631--1649, 2008.

\bibitem{lee1999lpo}
D.D. Lee and H.S. Seung.
\newblock {Learning the parts of objects by non-negative matrix factorization}.
\newblock {\em Nature}, 401(6755):788--791, 1999.

\bibitem{liu2005aoc}
I.~Liu and A.~Agresti.
\newblock {The analysis of ordered categorical data: an overview and a survey
  of recent developments}.
\newblock {\em TEST}, 14(1):1--73, 2005.

\bibitem{mare1980social}
R.D. Mare.
\newblock Social background and school continuation decisions.
\newblock {\em Journal of the American Statistical Association}, pages
  295--305, 1980.

\bibitem{mccullagh1980rmo}
P.~McCullagh.
\newblock {Regression models for ordinal data}.
\newblock {\em Journal of the Royal Statistical Society. Series B
  (Methodological)}, pages 109--142, 1980.

\bibitem{mohamed2010phone}
A.R. Mohamed and G.~Hinton.
\newblock {Phone recognition using restricted Boltzmann machines}.
\newblock In {\em Acoustics Speech and Signal Processing (ICASSP), 2010 IEEE
  International Conference on}, pages 4354--4357. IEEE, 2010.

\bibitem{paquet2011hierarchical}
U.~Paquet, B.~Thomson, and O.~Winther.
\newblock A hierarchical model for ordinal matrix factorization.
\newblock {\em Statistics and Computing}, pages 1--13, 2011.

\bibitem{podani2005multivariate}
J.~Podani.
\newblock Multivariate exploratory analysis of ordinal data in ecology:
  pitfalls, problems and solutions.
\newblock {\em Journal of Vegetation Science}, 16(5):497--510, 2005.

\bibitem{ranzato2010modeling}
M.A. Ranzato and G.E. Hinton.
\newblock {Modeling pixel means and covariances using factorized third-order
  Boltzmann machines}.
\newblock In {\em CVPR}, pages 2551--2558. IEEE, 2010.

\bibitem{rennie2005fmm}
J.D.M. Rennie and N.~Srebro.
\newblock {Fast maximum margin matrix factorization for collaborative
  prediction}.
\newblock In {\em Proceedings of the 22nd International Conference on Machine
  Learning (ICML)}, pages 713--719, Bonn, Germany, 2005.

\bibitem{robert1995simulation}
C.P. Robert.
\newblock Simulation of truncated normal variables.
\newblock {\em Statistics and computing}, 5(2):121--125, 1995.

\bibitem{salakhutdinov2009deep}
R.~Salakhutdinov and G.~Hinton.
\newblock {Deep Boltzmann Machines}.
\newblock In {\em Proceedings of The Twelfth International Conference on
  Artificial Intelligence and Statistics (AISTATS'09}, volume~5, pages
  448--455, 2009.

\bibitem{salakhutdinov2009replicated}
R.~Salakhutdinov and G.~Hinton.
\newblock {Replicated softmax: an undirected topic model}.
\newblock {\em Advances in Neural Information Processing Systems}, 22, 2009.

\bibitem{salakhutdinov2009semantic}
R.~Salakhutdinov and G.~Hinton.
\newblock Semantic hashing.
\newblock {\em International Journal of Approximate Reasoning}, 50(7):969--978,
  2009.

\bibitem{salakhutdinov2008probabilistic}
R.~Salakhutdinov and A.~Mnih.
\newblock {Probabilistic matrix factorization}.
\newblock {\em Advances in neural information processing systems},
  20:1257--1264, 2008.

\bibitem{Salakhutdinov-et-alICML07}
R.~Salakhutdinov, A.~Mnih, and G.~Hinton.
\newblock Restricted {B}oltzmann machines for collaborative filtering.
\newblock In {\em Proceedings of the 24th International Conference on Machine
  Learning (ICML)}, pages 791--798, 2007.

\bibitem{smolensky1986information}
P.~Smolensky.
\newblock {Information processing in dynamical systems: Foundations of harmony
  theory}.
\newblock {\em Parallel distributed processing: Explorations in the
  microstructure of cognition}, 1:194--281, 1986.

\bibitem{tieleman2008training}
T.~Tieleman.
\newblock {Training restricted Boltzmann machines using approximations to the
  likelihood gradient}.
\newblock In {\em Proceedings of the 25th international conference on Machine
  learning}, pages 1064--1071. ACM, 2008.

\bibitem{Truyen:2011b}
T.~Tran, D.Q. Phung, and S.~Venkatesh.
\newblock {Mixed-variate restricted Boltzmann machines}.
\newblock In {\em Proc. of 3rd Asian Conference on Machine Learning (ACML)},
  Taoyuan, Taiwan, 2011.

\bibitem{Truyen:2012b}
T.~Tran, D.Q. Phung, and S.~Venkatesh.
\newblock Sequential decision approach to ordinal preferences in recommender
  systems.
\newblock In {\em Proc. of the 26th AAAI Conference}, Toronto, Ontario, Canada,
  2012.

\bibitem{Truyen:2009a}
T.T. Truyen, D.Q. Phung, and S.~Venkatesh.
\newblock {Ordinal Boltzmann machines for collaborative filtering}.
\newblock In {\em Twenty-Fifth Conference on Uncertainty in Artificial
  Intelligence (UAI)}, Montreal, Canada, June 2009.

\bibitem{van2008visualizing}
L.~van~der Maaten and G.~Hinton.
\newblock {Visualizing data using t-SNE}.
\newblock {\em Journal of Machine Learning Research}, 9(2579-2605):85, 2008.

\bibitem{younes1989parametric}
L.~Younes.
\newblock {Parametric inference for imperfectly observed Gibbsian fields}.
\newblock {\em Probability Theory and Related Fields}, 82(4):625--645, 1989.

\end{thebibliography}
\end{document}